\documentclass{article}

\usepackage[margin=1in]{geometry}

\usepackage{apacite} 
\usepackage{graphicx}
\usepackage{url}
\usepackage{authblk}
\usepackage{booktabs}
\usepackage{stmaryrd}
\usepackage{turnstile}
\usepackage{centernot}
\usepackage{qtree} 
\usepackage{amsthm,amssymb}
\usepackage{mathtools}
\usepackage{subcaption}

\DeclarePairedDelimiterX{\infdivx}[2]{(}{)}{%
  #1\;\delimsize\|\;#2%
}

\newtheorem{claim}{Claim}
\newtheorem{universal}{Universal}

\newcommand{\tup}[1]{\left\langle #1 \right\rangle}

\newcommand{\eval}[2]{\left\llbracket #1 \right\rrbracket^{#2}}
\newcommand{\quantcomp}[0]{\mathcal{C}_Q^c}
\newcommand{\cons}[1]{cons$_{#1}$}
\newcommand{\lang}[0]{\mathcal{L}}

\begin{document}

\title{Compositionality and the lexicon in evolutionary semantics}
\author[*1]{Fausto Carcassi}
\affil[*]{Corresponding Author: f.carcassi@uva.nl}
\affil[1]{Institute for Logic, Language, and Computation, University of Amsterdam, Amsterdam, Netherlands}


\maketitle

\begin{abstract}
Formal semantics has shown that sentence meanings arise by recursively composing lexical meanings, yet much of the literature on semantic universals models either lexicons with fixed signal structures or holistic composition without interpretable lexical parts. We introduce a framework that integrates this fundamental insight of formal semantics in evolutionary modeling, by allowing lexical meanings and a composition function to co-evolve under pressures for conceptual simplicity and communicative accuracy. We apply this framework to the evolution of quantificational meaning. Analyzing the Pareto frontier, we find that the most well-known semantic universal, conservativity, emerges as an efficient system-wide abstraction. The account helps reconcile tensions between empirical evidence on quantifier learnability and prior evolutionary models. More broadly, the results demonstrate that the picture of sentential meaning developed in formal semantics can be productively combined with evolutionary modeling. The framework offers a template for studying universals that involve global compression within a grammatical category, semantic specialization of word classes, and the co-evolution of lexical and compositional meaning.
\end{abstract}

\vspace{0.5em}
\textbf{Keywords:} Conservativity $|$ Compositional Semantics $|$ Language Evolution $|$ Semantic Universals

\vspace{0.5em}

\section{Introduction}

Compositionality is one of the fundamental features of human language. As stated in \citeA{parteeCompositionality1984}, this principle says that ``The meaning of an expression is a function of the meanings of its parts and of the way they are syntactically combined.'' In the past 60 years, a sophisticated picture of compositional meaning has been developed \cite{aloniCambridgeHandbookFormal2016,heimSemanticsGenerativeGrammar1998,dowtyIntroductionMontagueSemantics1981,montagueProperTreatmentQuantification1973,lewisGeneralSemantics1970}. While details vary, the wide consensus is that the literal meaning of a sentence is constructed by recursively composing the meaning of its syntactic constituents, starting from the interpretation of lexical leaves up to the full sentence. Natural language semantics then has two main ingredients: the lexical meanings and the operation that composes them. The interaction of these two ingredients is taken to be the basis for the expressive power of natural language.

The formalization of lexical semantics made it then possible to describe many \textit{semantic universals}: universal features of the meaning of lexical entries attested in languages across the world. Such universals have been identified in disparate semantic domains, most famously quantification \cite{barwiseGeneralizedQuantifiersNatural1981,keenanSemanticCharacterizationNatural1986}.\footnote{Other domains studied in recent literature include kinship \cite{kempKinshipCategoriesLanguages2012,mollicaLogicalWordLearning2021}, color \cite{zaslavskyEvolutionColorNaming2022,twomeyWhatWeTalk2021,chaabouniCommunicatingArtificialNeural2021,steinert-threlkeldEaseLearningExplains2020,zaslavskyEfficientCompressionColor2018,gibsonColorNamingLanguages2017}, number \cite{xuNumeralSystemsLanguages2020,denicRecursiveNumeralSystems2024}, artifacts/containers \cite{zaslavskySemanticCategoriesArtifacts2019}, spatial language \cite{chenInformationtheoreticApproachTypology2023}, Boolean connectives \cite{bar-levCommunicativeStabilityTypology2025,uegakiInformativenessComplexityTradeoff2024,carcassiAssertionDenialEvolution2023,icardSimpleLogicConcepts2023,sbardoliniLogicLexicalConnectives2023,hornSemanticPropertiesLogical1972}, indefinite pronouns \cite{denicComplexityInformativenessTradeoff2021}, modals \cite{imelModalSemanticUniversals2022}, and gradable adjectives \cite{carcassiCulturalEvolutionScalar2020,carcassiEvolutionAdjectivalMonotonicity2019}.}
The field of evolutionary semantics has then developed explanations for these universals, showing that in many cases attested inventories optimize for simplicity and/or usefulness for communication (\citeA{steinert-threlkeldEaseLearningExplains2020,carcassiMonotoneQuantifiersEmerge2021}, and \citeA{vandepolQuantifiersSatisfyingSemantic2023} for quantifiers. See \citeA{kempSemanticTypologyEfficient2018,gibsonHowEfficiencyShapes2019} for an overview). While the lexical meanings discussed in this work could in principle participate in complex semantic composition, most models only look at collections of unstructured signals.\footnote{Another important strand of research in evolutionary semantics focuses on the overall semantic organization of language, with a special focus on the evolution of compositional structure \cite{chaabouniCompositionalityGeneralizationEmergent2020,mordatchEmergenceGroundedCompositional2018a,kirbyCompressionCommunicationCultural2015,smithIteratedLearningFramework2003,kirbySpontaneousEvolutionLinguistic2001,nowakEvolutionLanguage1999}. Other work in this line focuses on the evolution of signaling systems and their emergent features, such as ambiguity or categorization \cite{skyrmsSignalsEvolutionLearning2010a,lewisConventionPhilosophicalStudy2002}. This literature usually defines a graded measure of compositionality based on the mapping from signal space to meaning space (e.g., topographic similarity \cite{brightonUnderstandingLinguisticEvolution2006} and positional disentanglement \cite{lazaridouEmergenceLinguisticCommunication2018}). As discussed in \citeA{andreasMeasuringCompositionalityRepresentation2019}, deep learning agents present the additional problem of reconstructing the lexicon and composition function. However, these models typically lack individually defined, interpretable lexical meanings composing recursively like in natural language.}

Beyond lexical universals, previous work in evolutionary semantics has also studied the evolution of compositionality.
Several papers have looked at the evolution of compositionality in Lewisian signaling games \cite{skyrmsSignalsEvolutionLearning2010a,lewisConventionPhilosophicalStudy2002,barrettEvolutionCodingSignaling2009,frankeEvolutionCompositionalitySignaling2016}. \citeA{steinert-threlkeldEmergenceNontrivialCompositionality2020} points out that in much of this work the focus is on the emergence of compositionality \textit{as such}, and therefore the specific composition functions that evolve are generally only encoding conjunction of meanings, which he calls \textit{trivial composition} (See also \citeA{barrettEvolutionCompositionalLanguage2020} for a more recent version of this approach). In turn, \citeA{steinert-threlkeldEmergenceNontrivialCompositionality2020} develops a model of the evolution of a system encoding items that compose by function application. In contrast to this work, in this paper we assume compositionality and instead focus on how the tradeoff in complexity between compositionality and the lexicon in the context of specific semantic universals.

In this paper, we argue that methodological differences between formal and evolutionary semantics have made it difficult to use the characterization of the universals coming from the former to develop explanations in the latter. We consider \textit{quantificational conservativity}, arguably the best-known semantic universal \cite{barwiseGeneralizedQuantifiersNatural1981}, as our case study. We discuss several alternative characterizations of this universal, and show that a novel characterization that involves the interaction of lexical meanings and composition is better suited for evolutionary semantics. We then present a novel computational framework that models both lexical meanings and an explicit composition function, to find systems that strike a tradeoff between evolutionary pressures. We show that our evolutionary model of the evolution of conservativity reconciles tensions between the data and previous accounts.

The core of our account is that, rather than a feature of the meaning of individual quantifiers, conservativity is part of how languages \textit{compose} determiners with their arguments. Since it needs to be defined only once in the composition function, conservativity only requires a small increase in the global complexity of a language. On the other hand, since it impacts the interpretation of all quantified sentences at once, it can have a large impact on communicative accuracy. We show that languages striking the best compromise between complexity and communicative accuracy use the conservativity abstraction. Moreover, we show that this result crucially depends on the interaction between the meaning of signals, their pragmatic enrichments, and the language's global complexity.

This paper makes three contributions. First, it proposes a model of how the syntax-semantics interface and \textit{system-level} complexity play a role in language evolution, via high-level abstractions in the composition function. Second, it provides a novel account of the universal of conservativity. Lastly and most importantly, it studies evolutionary tradeoffs in a model of compositional meaning that strongly resembles the picture of modern formal semantics. In this way, it further develops the bridge between evolutionary models of linguistic universals and modern formal analyses of meaning.

The paper is structured as follows. In Section \ref{sec:conservativity}, we introduce the running case study of conservativity and consider several related characterizations of the universal. In Section \ref{sec:categorematic}, we discuss several ways of characterizing semantic universals and why they are importantly different in the context of evolutionary semantics, using conservativity as an example. In Section \ref{sec:framework}, we describe the framework that we use to study the tradeoff of lexical meaning and composition. In Section \ref{sec:previouslit}, we discuss previous empirical and theoretical work on conservativity, including previous evolutionary explanations and their weaknesses. In Section \ref{sec:tradeoffmodel}, we present our tradeoff model for quantification, and in Section \ref{sec:results} the results. Finally, we discuss the implications of the results in Section \ref{sec:discussion} and conclude in Section \ref{sec:conclusions}.

\section{A case study: Quantificational conservativity}\label{sec:conservativity}

In this section, we introduce the semantic universal that we will use to demonstrate the framework and make it more concrete, namely \textit{conservativity}. This is a universal of the meaning of lexical determiners, e.g., `some', `all', `three', `most' in English. These are grammatically simple expressions that take a common noun as argument (its \textit{first}, \textit{left} or \textit{internal} argument) and return a determiner phrase, e.g., `three cats', which in turn can take a verb phrase (its \textit{second}, \textit{right} or \textit{external} argument) to produce a sentence, e.g., `three cats sleep'. The tradition of Generalized Quantification Theory sees natural language determiners as expressing relations between sets, i.e., semantic objects of type $\tup{\tup{e,t},\tup{\tup{e,t},t}}$ \cite{barwiseGeneralizedQuantifiersNatural1981,keenanSemanticCharacterizationNatural1986}. Any semantic object of that type in turn is a possible meaning for a natural language determiner. Various universals of quantification have been proposed that seem to constrain the typological variation in the meaning of attested determiners, the most famous of these arguably being \textit{conservativity}.

Slightly generalizing the definitions in \citeA{zuberNoteConservativity2019}, we call a determiner $Q$ \textit{conservative} on its $n$th argument, or \textit{cons$_n$}, iff in any occurrence replacing its $n$th argument with its intersection with its other arguments preserves its truth value.\footnote{Formally, on every universe $U$ and $P_1, \ldots, P_N \subseteq U$:
\begin{equation*}
Q_U(P_1, \ldots, P_{n-1}, P_n, \ldots, P_N) \iff Q_U(P_1, \ldots, P_{n-1}, \bigcap_{i=1}^N P_i, \ldots, P_N)
\end{equation*}
We will omit the dependence on $U$ when not relevant. 
Alternatively, we could define \cons{n} as saying that a quantifier Q \textit{lives on an argument}, i.e. $Q(P_1, \ldots, P_n, \ldots, P_N) = Q(P_1 \cap P_n, \ldots, P_n, \ldots, P_N \cap P_n) $. This is arguably a more intuitive definition, but we choose the definition in the main text for consistency with previous literature. Since we will only consider quantifiers with two arguments, the distinction will not matter for the discussions in this paper.}
Assuming $Q$ has two arguments, conservativity on the second argument says that the truth of $Q(L, R)$ does not depend on anything outside the first argument $L$. For instance, because the English determiner `all' is conservative, `all penguins walk' is true exactly when `all penguins are penguins that walk'.

Much work has claimed that \cite{barwiseGeneralizedQuantifiersNatural1981,keenanSemanticCharacterizationNatural1986}:\footnote{\citeA{keenanSemanticCharacterizationNatural1986} specifically claim that not only the syntactically simple, but also the complex determiners are conservative. This is in contrast to other universals of quantification such as \textit{monotonicity}, which only apply to simple lexicalized determiners. We assume this for the rest of the paper. While we do not specifically study complex determiners, this observation is particularly interesting in light of the fact that, as will become clear below, conservativity in our model is a feature of any natural language expression of quantifier type rather than just the simple ones.}
\begin{universal}\label{claim:cons2det}
    All natural language determiners express \cons{2} quantifiers. 
\end{universal}
\noindent For the purpose of this paper, we consider a slightly different generalization stated in terms of semantic type, for reasons that will become clear below:
\begin{universal}\label{claim:cons2}
    All natural language expressions denoting relations between two sets (i.e., meanings of type $\tup{\tup{e,t}, \tup{\tup{e,t},t}}$) express \cons{2} quantifiers. 
\end{universal}
\noindent In most accounts, these two universals are equivalent, because only determiners express relations between two sets. We assume that this is the case below and speak interchangeably of `determiners' and terms that express relations between two sets.

Conservativity is a remarkable generalization about human language, for three reasons. First, it poses an extremely strong constraint on possible quantifier meaning, massively reducing the space of possible quantifiers. Second, some of these non-conservative quantifiers are seemingly simple. For instance, a hypothetical quantifier `equi', expressing that two sets have the same size, would not be conservative, because `equi dogs sleep' is not true if and only if `equi dogs are sleeping dogs' (e.g., an awake dog falsifies the latter but not the former). Third, the generalization does not extend to at least superficially similar cases in the verbal domain, such as `outnumber' (which takes plural individuals as arguments). At a conceptual level, `outnumber' can be naturally thought of as expressing a non-conservative meaning \cite{knowltonStrengthConservativityEvidence2025}.

An apparent counterexample to Universal \ref{claim:cons2} is the word `only'. `Only' seems to express a non conservative quantifier because, for instance, the truth of the utterance `only dogs bark' depends on entities that are not in the set of dogs.\footnote{Other proposed counterexamples include `mostly' and `just'.} A possible answer is to deny that `only' expresses a relation between sets, and that therefore it falls outside of the scope of Universal \ref{claim:cons2}:
\begin{claim}\label{claim:only}
    Some apparent exceptions to conservativity, such as `only', do not in fact express relations between two sets.
\end{claim}
\noindent Claim \ref{claim:only} is supported by a substantial amount of previous literature that takes it to be cross-categorial rather than targeting sets \cite{rooth_association_1985,rooth_theory_1992,von_stechow_current_1990}.

A second option is to slightly weaken the universal itself to include `only,' as \citeA{zuber_generalising_2010} does, introducing the notion of \textit{weak conservativity} (see also \citeA{zuberNoteConservativity2019} and \citeA{iemhoff_weak_2019}):
\begin{universal}\label{claim:weakcons}
    All simple determiners express \cons{1} or \cons{2} quantifiers.
\end{universal}
\noindent `Only' satisfies this weakened requirement because, e.g., `only cats sleep' entails that `only sleeping cats sleep'.

From an evolutionary point of view, Universal~\ref{claim:cons2} contains two conceptually distinct claims about conservativity. First, determiners within each language express quantifiers that are uniformly conservative on a specific argument, be it the first or the second, but possibly different across languages. Second, across languages the conservative argument is always the second, rather than the first. For reasons that will become clear below, these two claims might need separate explanations and we should keep them distinct. Therefore, while conservativity is traditionally formulated in terms of individual determiners, the possibility of conservativity on different arguments suggests an additional universal at the level of systems of determiners in a language. We call a \textit{set} of determiners \cons{n} iff all of its members are \cons{n}. We define the following universal:
\begin{universal}\label{claim:globalcons}
    For each natural language, there is an $n$ such that the set of all its determiners is \cons{n}.
\end{universal}
\noindent Universal \ref{claim:globalcons} is weaker than Universal \ref{claim:cons2}, since it leaves open the possibility of a language with \cons{1} determiners, but it is stronger than Universal \ref{claim:weakcons}, since it excludes the possibility of mixing in a single language quantifiers that are \cons{1} (and not \cons{2}) with quantifiers that are \cons{2} (and not \cons{1}).

We assume that the purported counter-examples to Universal \ref{claim:cons2} in fact do not express quantifiers, and therefore that Universal \ref{claim:cons2} is empirically adequate. We introduce Universal \ref{claim:globalcons} not because there are empirical counterexamples to Universal \ref{claim:cons2}, but rather because we seek an explanatory account for the former that does not necessarily determine with respect to which $n$ the determiners are conservative. Other explanations for the selection of the arguments are available based, e.g., on information structure and processing \cite{lidzInterfaceTransparencyPsychosemantics2011}.

For reasons of space, the discussion of conservativity in this paper sets aside various complications. First, while conservativity is taken to be universal, quantification in natural language shows interesting typological variation in many dimensions \cite{bach_quantification_1995,keenan_handbook_2012}. Second, there are grammatical categories beyond determiners whose members denote quantifiers, so-called A-quantification, e.g., `always', `sometimes', `generally' \cite{lewis_adverbs_1975,partee_quantificational_1995}. While these could in principle provide useful data, their analysis is more controversial.\footnote{It is worth noting here that these have been argued to be quantifiers over a different types, see e.g., \citeA{de_swart_adverbs_1991}, so that much of the discussion below could apply to them.} We leave it to future work to explore how these fit into the picture below.

\section{Types of semantic analysis in formal and evolutionary semantics}\label{sec:categorematic}

In this section, we argue that a given semantic universal can be characterized in multiple ways that are equivalent for formal semantics but importantly different for evolutionary semantics, using conservativity as an example. In contemporary analyses, a semantic system is typically specified as a combination of a \textit{lexical interpretation function}, which assigns an interpretation to each lexical entry, and a composition function consisting of a set of \textit{modes of composition}, i.e., rules describing how meanings compose recursively. The two are related by rules that determine which mode of composition is involved in particular instances of composition. The most common approach, which we assume in the rest of this paper, is \textit{type-driven composition}, which lets the semantic type of the involved constituents determine the mode of composition \cite{klein_type-driven_1985}. Interpretation function and composition modes together should correctly account for the interpretation of whole sentences, but different theories vary in what semantic content they place in the lexical entries and what in the composition function.

We can distinguish two general strategies. On the one hand, the \textit{categorematic} strategy places all the semantic content in the lexical entry, and gives an analysis of the composition function that is as general as possible. This is the strategy favored in mainstream compositional formal semantics, exemplified by \citeA{heimSemanticsGenerativeGrammar1998}. For instance, a categorematic description of quantifiers would define function application as a generic composition function, and place all the semantic content in the interpretation of the determiners:\footnote{Assuming that $\alpha : \tup{\tau, \upsilon}$ and $\beta : \tau$. We follow the standard notation of writing $f : \tau$ to indicate that object $f$ is of type $\tau$. We use $\tau$ and $\upsilon$ as variables over types, and $e$, $t$, and $s$ as the basic types of individuals, truth values, and possible worlds respectively.}
\begin{align}\label{eq:quantcat}
    \eval{ \alpha \beta }{} &= \eval{\alpha}{} ( \eval{\beta}{} )\\
    \eval{\text{every}}{} &= \lambda L. \lambda R. |L\cap R|=|L|\nonumber\\
    \eval{\text{some}}{} &= \lambda L. \lambda R. |L\cap R| > 0 \nonumber\\
    \eval{\text{no}}{} &= \lambda L. \lambda R. |L\cap R| = 0 \nonumber
\end{align}
This analysis has a simple, generic composition function that functions in the same way across determiners and other grammatical categories. Categorematic analyses are a practical choice to keep the semantics-syntax interface transparent and making it possible to easily accommodate new lexical entries.

On the other hand, the \textit{syncategorematic} strategy analyses lexical entries as essentially semantically empty in isolation, and instead encodes meaning by directly specifying how they compose with other expressions. For example, this would be a syncategorematic analysis for the same determiners:\footnote{When $f$ is a characteristic function, we write $| f |$ for the number of elements that verify it.}
\begin{align}\label{eq:quantsyncat}
    \eval{[[\text{every}\, \alpha_{\text{NP}} ] \, \beta_{\text{NP}}]}{} = 1 &\iff |\eval{\alpha}{} \cap \eval{\beta}{}| = |\eval{\alpha}{}| \\
    \eval{[[\text{some}\, \alpha_{\text{NP}} ] \, \beta_{\text{NP}}]}{} = 1 &\iff |\eval{\alpha}{} \cap \eval{\beta}{}| > 0 \nonumber\\
    \eval{[[\text{no}\, \alpha_{\text{NP}} ] \, \beta_{\text{NP}}]}{} = 1 &\iff | \eval{\alpha}{} \cap \eval{\beta}{}| = 0 \nonumber
\end{align}
Note that this analysis does not directly assign a meaning to the quantifiers in isolation, but rather specifies a separate mode of composition for each quantifier, in a certain syntactic construction. Syncategorematic analyses are practical when a clear set of logical terms is fixed in advance, e.g., for model-theoretical semantics of logical systems. A syncategorematic analysis of natural language can also be motivated by non-denotational accounts of meanings, e.g., inferentialism \cite{brandomArticulatingReasonsIntroduction2003}. Typically, some part of the lexicon is still analyzed categorematically, e.g., content words like nouns and verbs. The variety of syntactic contexts in which a single word appears in natural language can make syncategorematic definitions more complex.%
\footnote{Sometimes the term `syncategorematic' is used for analyses where each syntactic construction gets associated with a separate mode of composition, e.g., the several rules of function application in \citeA{montagueProperTreatmentQuantification1973}.}

In the case of the system of quantifiers, the most compact representation is somewhere in between. Both the analyses in Eq.~\ref{eq:quantcat} and Eq.~\ref{eq:quantsyncat} contain the repeated expression $L \cap R$. This expression could be encoded just once as part of how determiners are interpreted as a class. This simplifies the whole system:
\begin{align}\label{eq:quantsemicat}
    \eval{\gamma \alpha}{} &= \lambda \beta. \eval{\gamma}{}(\eval{\alpha}{})(\eval{\alpha}{} \cap \eval{\beta}{})  &\text{if $\eval{\gamma}{} : \tup{\tup{e,t}, \tup{\tup{e,t}, t}}$}\\
    \eval{\text{every}}{} &= \lambda L. \lambda R. |R|=|L|\nonumber\\
    \eval{\text{some}}{} &= \lambda L. \lambda R. |R| > 0 \nonumber\\
    \eval{\text{no}}{} &= \lambda L. \lambda R. |R| = 0 \nonumber
\end{align}
Effectively, this system is saying that the right argument to the determiner is always restricted by its left argument. As we will see below, this corresponds to the generalization of \textit{conservativity} in quantifiers. The gap in complexity between this analysis and the two analyses above increases as more quantifiers are added.\footnote{A feature of the analysis of Eq.~\ref{eq:quantsemicat} is that the interpretation function explicitly conditions on the semantic types of its arguments. This raises the question of whether composition can be sensitive to features other than argument type. The answer to this question determines what analysis is available. For example, determiners can be categorized into proportional and cardinal, corresponding to two semantic schemes:
\begin{align*}
    \eval{Q_p}{} &= \lambda L. \lambda R. |L \cap R| \cdot |L| &\text{proportional} \\
    \eval{Q_c}{} &= \lambda L. \lambda R. |L \cap R| \cdot N &\text{cardinal}
\end{align*}
where $\cdot \in \{ =, >, < \}$ and $N \in \{0, 1\}$. These two schemes could be encoded in the composition function, assuming that composition is sensitive to arbitrary subclasses of determiners.
While a detailed study of these issues is beyond the scope of this paper, previous literature has shown that the mode of composition can depend \textit{at least} on the involved types. For instance, if $\alpha$ and $\beta$ are both of type $\tup{e,t}$, $\eval{\alpha \beta}{}$ is composed by \textit{predicate modification}, essentially intersection for characteristic functions \cite{heimSemanticsGenerativeGrammar1998}.%
}

We call the type of analysis in Eq.~\ref{eq:quantsemicat} \textit{semi-categorematic}, since it includes a categorematic component that assigns independent semantic content to the lexical entries, and a syncategorematic component, which substantially adds to the semantic content based on the syntactic context. In the case of determiners, the semi-categorematic analysis is simpler than the preceding two: it encodes the same information with a shorter description \cite{grunwaldMinimumDescriptionLength2007}. The exact measure of complexity and the choice of metalanguage can influence what the simplest analysis is. 
In this paper, we make some provisional intuitive choices that can be further supported in future literature.
In general, semantic content shared by several lexical entries $C$ is best analyzed semi-categorematically as part of how elements of $C$ compose. This happens for instance whenever arguments are semantically specialized across a word class.\footnote{There are other ways to compress the semantic system other than redistributing semantic content between lexicon and composition function. For instance, auxiliary variables could be defined that are reused across the lexicon, e.g.:
\begin{align*}
    \eval{ \alpha \beta }{} &= \eval{\alpha}{} ( \eval{\beta}{} )  &\text{Generic composition function}\\
    \texttt{let } f &= \lambda L, R. | L \cap R | &\text{Conceptual abstraction}\\ 
    \eval{\text{every}}{} &= \lambda L. \lambda R. f(L,R) = |L| &\text{Lexical entry}
\end{align*}
However, for the applications we consider below, the need to manage arguments makes this kind of system more complex than the semi-categorematic ones discussed above.}

The choice between these styles of analysis in formal semantics is mostly seen as a matter of flexibility and ease of description, rather than a substantial commitment. This is because the main goals of formal semantics, i.e., a characterization of logical form, truth-conditions, and inference patterns, can be achieved equally well in these styles of analysis. However, the difference in descriptive complexity is not typically essential from the perspective of formal semantics. On the other hand, differences in descriptive complexity are crucial for evolutionary semantics, where explanations of semantic universals often appeal to representational complexity, arguing that universals strike a balance between simplicity and communicative accuracy (for discussions of the relation between compositionality in general and complexity see, e.g., \citeA{pagin_communication_2012}). It is hard to determine in advance which style of analysis offers the simplest characterization of a given universal. Therefore, it is important in evolutionary semantics to use a descriptive language that can capture different characterizations, which might be categorematic, syncategorematic, or semi-categorematic.

Descriptive complexity also matters in evolutionary explanation when the formal analysis is taken to represent the cognitive encoding of the semantic system. This has been done for instance in the Language of Thought tradition in cognitive science \cite{fodorLanguageThought1975,piantadosiFourProblemsSolved2016}. In this case, the metalanguage is assumed to have a cognitive reality, and the choice of operators can matter in empirical predictions of, e.g., learnability. This is the approach we take in the model below. This approach raises the question of which operators should be used for a specific domain. While the selection of operators in the Language of Thought that best captures human learning is not known,\footnote{But see \citeA{piantadosiBootstrappingLanguageThought2012} for an attempt at recovering such an inventory, and \citeA{carcassiBooleanLanguageThought2023} for a methodological study of recoverability in the Boolean domain.} in the model below we consider a simple selection of primitives that does not advantage conservativity (see Appendix~\ref{appendix:prior} for quantitative assessments of the prior advantage of conservativity with the chosen LoT).

In this section, we have discussed two traditional ways of encoding the same overall semantic content, and we have argued that the choice between the two in evolutionary semantics should be guided by descriptive simplicity. The point is not simply that complexity depends on the chosen representational language, the details of which can depend on arbitrary and provisionally partially motivated choices both in formal and in evolutionary semantics, but rather that the shape of the analysis is motivated by different considerations.\footnote{We thank an anonymous reviewer for pointing out the need to clarify this point.} In this context, we have shown that analyzing a semantic domain with the goal of minimizing descriptive complexity can lead to a semi-categorematic analysis, which distributes the semantic content between the individual lexical entries and the composition function. Determiners are one example where a semi-categorematic analysis is simpler than traditional ones, and we leave an exploration of other cases to future work.

\section{A framework for tradeoff analysis in a compositional language}\label{sec:framework}

In this section, we sketch a computational framework for tradeoff analysis flexible enough to represent different styles of analysis as discussed in the previous section. This framework broadly implements a view of communication that is standard in formal semantics. The framework includes a communication game involving two agents that pragmatically produce and interpret signals that restrict the common ground. The language consists of an interpreted lexicon and a composition function. Based on the discussion above, this model has the resources to distribute semantic content between these two in order to minimize the overall complexity. The evolution of a portion of the language (e.g., the meaning of its nouns, quantifiers, composition function, or some combination thereof) can be studied by leaving it unspecified, defining a space of possible options via a formal grammar, and evaluating them with respect to various evolutionary pressures. More technical details concerning this Section are given in Appendix \ref{appendix:technicalDetail}.

\subsection{The model of communication and interpretation}\label{sec:model}

The world consists of a sender $S$, a receiver $R$, and a context $c$ consisting of five objects. Each object has a single feature, namely an integer between -10 and 10, which both $S$ and $R$ can see, and is either a \textit{target} or a \textit{distractor}, which only $S$ can see. Each communicative game unfolds in three stages:
\begin{enumerate}
    \item A context $c$ is sampled.
    \item $S$ constructs a signal $s$ that expresses a proposition, i.e., a function from a context to a truth value. 
    \item $R$ uses $s$ to update their beliefs about which objects are targets.
\end{enumerate}
The success of the communicative exchange is defined by how few possible contexts remain after $R$ has updated the common ground. The standard model of communication in evolutionary semantics is based on \textit{signaling games}, where a sender observes one of a finite set of states and sends a signal to a receiver, who then guesses the state \cite{lewisConventionPhilosophicalStudy2002}. If the observed state and the guessed state are the same, the communication is successful. The current model departs from this setup in that the agents share partial information about the world (their \textit{common ground}), communication aims at describing the world rather than discriminating between a handful of scenes, and signals are compositionally structured. The resulting picture of communication is more similar to the one in formal semantics, helping build a more explicit bridge between the latter and evolutionary modeling.

The agents' signals belong to a shared language. The language has two ingredients: a lexicon and an interpretation function that associates each signal with a meaning.\footnote{Importantly, the underlying implementation is easily extensible to include other types, meanings, and forms of composition functions.} The lexicon consists of a small but expressive set of strings, the `words' of the language, including `even', `prime', `and', `or', `first object', etc. Lexical entries can also be combined recursively in binary trees to construct syntactically complex signals. The syntax of the language is type-driven, and any two constituents can be combined together to form a larger constituent iff their meanings compose. 

The interpretation function takes any signal $\alpha$, which can be simple or complex, and returns its interpretation $\eval{\alpha}{c}$ in the context $c$. The meanings of the lexical entries are specified in advance except for the target of investigation in the case study below, namely the three quantifiers $Q_1$, $Q_2$, and $Q_3$. If the signal is complex, its interpretation is calculated by recursively composing its constituents. The composition operation is also defined explicitly as part of the interpretation function, which makes the language flexible enough to encode the semi-categorematic analyses discussed in Section~\ref{sec:categorematic}. By default, composition happens simply by function application. However, we treat separately the part of the composition function that composes a quantifier with its arguments, which we denote with $\quantcomp$, which we leave unspecified and free to vary as described in the next section.

$S$'s goal in the communicative game is to send a signal that allows $R$ to exclude as many contexts as possible, thereby increasing communicative success. Therefore, $S$'s signals always express a proposition, i.e., a function from contexts to truth values. With the lexicon and composition function of the language in this model, signals expressing propositions can be constructed in various ways, e.g., via a combination of a unary predicate with a proper noun, which only relies on function application for composition:
\vspace{0.2cm}
\begin{center}
\footnotesize
\Tree [.{$\eval{\text{target}}{c}(\eval{\text{2e}}{c})$} 
    {target\\$\eval{\text{target}}{c}$} 
    {2e\\$\eval{\text{2e}}{c}$} 
]
\end{center}
\vspace{0.2cm}
A more important case for this model is the case of a quantifier combined with two predicates, which composes using the function $\quantcomp$:
\vspace{0.2cm}
\begin{center}
\footnotesize
\Tree [.{$\quantcomp(\eval{\text{Q}_1}{c})(\eval{\text{target}}{c})(\eval{\text{even}}{c})$} 
    [.{$\quantcomp(\eval{\text{Q}_1}{c})(\eval{\text{target}}{c})$} 
        {Q_1\\$\eval{\text{Q}_1}{c}$} 
        {target\\$\eval{\text{target}}{c}$}  
    ]
    {even\\$\eval{\text{even}}{c}$}    
]
\end{center}
\vspace{0.2cm}
Since the space of signals cannot be searched exhaustively for computational reasons, when running a specific communication game the agents consider all well-formed signals up to a fixed depth of 3, which includes both of the examples above (of depth 2 and 3 respectively). The depth also provides a way to capture utterance cost, since with unlimited depth the agents could trivially form a conjunction specifying for each entity whether it is a target.\footnote{The implementation also includes a mechanism to include presuppositions in the lexical entries. Rather than evaluating the subtree to false, a presupposition failure interrupts composition and evaluates the whole signal it is part of as incompatible with the considered context. Since this mechanism is not relevant for the model at hand, we leave it out of the exposition in the main text.} 

Real communication involves enrichment of the literal meaning of utterances with additional pragmatic inferences. This is particularly well-studied for the case of quantifiers, where for instance an assertion of `some' tends to produce an implicature that `not all'.\footnote{We ignore questions of QUD since in the communication game of our simulation the QUD remains the same throughout. Future work can explore models with a varying QUD.} Therefore, $R$ enriches the literal interpretation of the signal $s$ with the negation of all the signals that would be strictly more informative than $s$ in the common ground. This is a simple model of Gricean implicatures. Previous work has shown that pragmatics can play an important role in the evolution of semantic structure  \cite{woensdregt_computational_2021,carcassiEvolutionAdjectivalMonotonicity2019,brochhagen_coevolution_2018}. More details on the implementation of pragmatic communication can be found in Appendix~\ref{subappendix:pragmatics}.

\subsection{Partially specified languages and tradeoff analysis}

In line with much previous literature, we aim to show that some languages satisfying the universal under consideration strike an optimal compromise between different pressures. This kind of tradeoff analysis therefore requires a space of possible languages that are evaluated with respect to the dimensions of optimization. In our framework, possible languages are obtained by leaving some component of the semantics free to vary. In the case study of conservativity, this is the meaning of some lexical entries and a part of the composition function, $\quantcomp$. We use a formal grammar to define the possible values for these components of the language, in line with previous work in evolutionary semantics \cite{kempKinshipCategoriesLanguages2012,carrSimplicityInformativenessSemantic2020,mollicaLogicalWordLearning2021}.

Within this space of possible languages, we focus on the tradeoff between two pressures. The first pressure is communicative utility, which languages tend to maximize. This is the expected utility of a communicative game played with that language. This measure of communicative accuracy includes two innovations. First, signals are compositionally structured. This plays a crucial role in the simulation since the language is specified not just by the lexical semantics but also by the composition function. Second, agents do not search a manually, pre-specified set of signals. Instead, the space of signals is induced generatively as any binary tree with leaves in the lexicon and constituents that can be composed with the given composition function. Agents search this space of signals up to a depth bound.

The second pressure is conceptual complexity, which languages tend to minimize. The complexity of a language is defined as the descriptive complexity of the interpretation function. In line with much previous work, we take this to track the cognitive complexity of the semantic system, and we measure this as the negative log probability of a derivation in a Probabilistic Context Free Grammar \cite{fodorLanguageThought1975,piantadosiFourProblemsSolved2016,piantadosiBootstrappingLanguageThought2012,piantadosiLogicalPrimitivesThought2016}, described in detail in Appendix~\ref{appendix:conservativity}. 
Our measure is encoded not at the level of single lexical entries, but rather as the \textit{global complexity} of the interpretation function, including the composition function. This allows abstractions defined in the composition function to be reused across the language. We will see that this plays a crucial role in predicting the emergence of conservativity. Moreover, we show that complexity alone, even when measured globally, is not enough to predict conservativity without a pressure for communicative accuracy. 

\begin{figure}[t]
    \centering
    \includegraphics[width=0.7\textwidth]{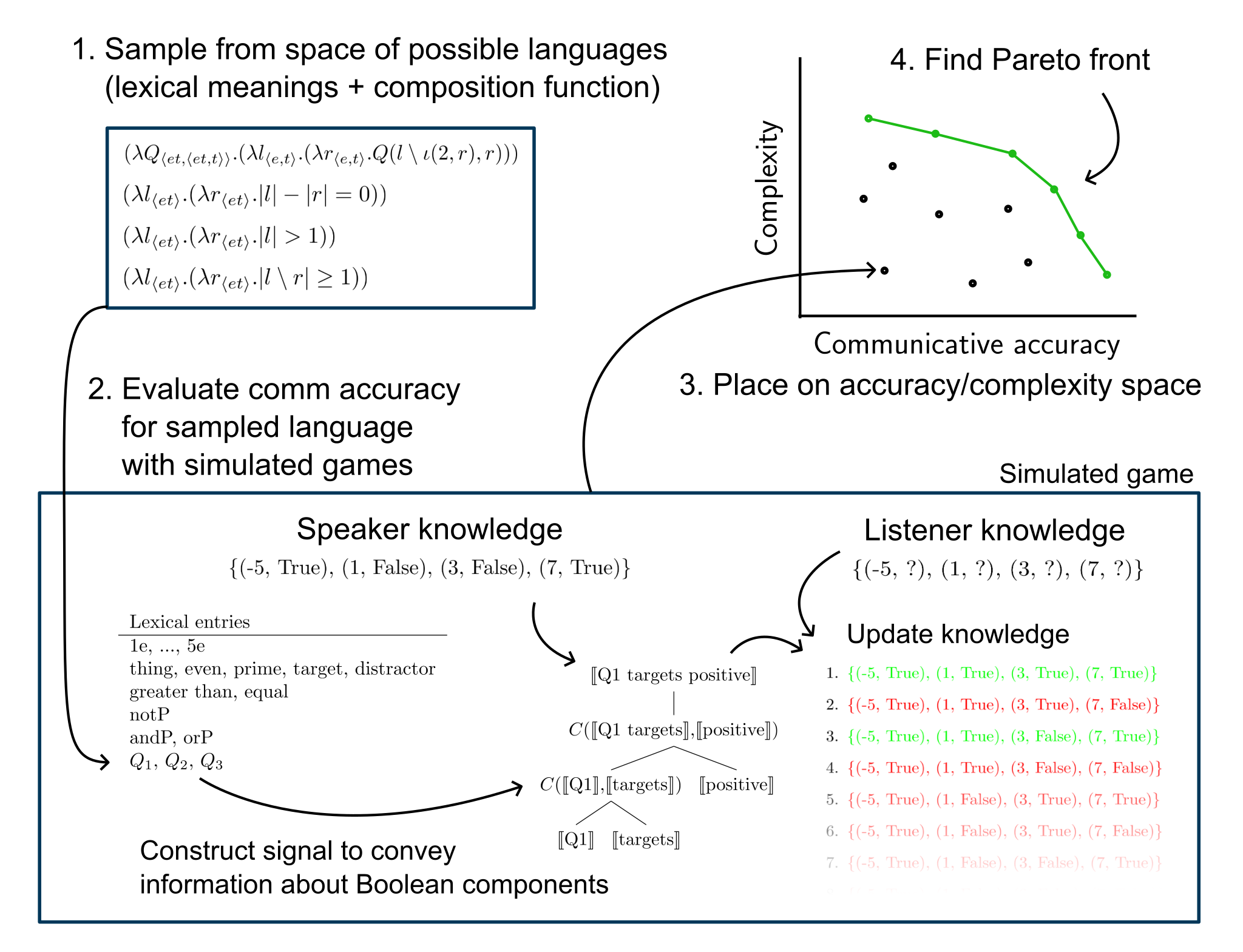}
    \caption{Diagram illustrating the overall structure of the model when doing a tradeoff analysis by leaving unspecified three quantifiers and the part of the composition function that composes quantifiers with their arguments.}
    \label{fig:modeldiagram}
\end{figure}

Figure \ref{fig:modeldiagram} displays the overall structure of the model for the case study. The model explores a space of systems of quantifiers to find the ones that strike the best tradeoff between the two pressures. A combination of quantifier meanings and composition function is sampled, then their communicative accuracy is evaluated by averaging over a number of simulated communication games. After repeating this procedure many times, we consider the Pareto frontier of the languages that strike the best compromise.

\section{A novel explanation for conservativity}

\subsection{Previous accounts and empirical work on conservativity}\label{sec:previouslit}

Before applying our framework to quantification, we review the empirical and theoretical landscape that emerged in previous literature, focusing on Universal \ref{claim:cons2}, since it has received much attention in the literature. Since Universal \ref{claim:cons2} is stronger than Universal \ref{claim:globalcons}, proposed explanations of the former would also be explanations of the latter.

Three important explanatory strategies are listed and discussed in \citeA{knowltonDeterminersAreConservative2020}. We report them briefly here. The first is \textit{lexical restriction}, articulated for instance in \citeA{keenanSemanticCharacterizationNatural1986}. This position stipulates that determiner meanings are fully constructed in a conceptual language that can only express conservative meanings. As pointed out by \citeA{knowltonDeterminersAreConservative2020}, this is arguably best characterized as a redescription of the explanandum rather than an explanation.

The second explanatory strategy is \textit{interface filtering}, articulated in \citeA{romoliStructuralAccountConservativity2015}. This argues that conservativity is a consequence of the syntax-semantics interface. Specifically, the quantificational DP moves up and leaves a trace that encodes the intersection of the internal and external sets, so that the second semantic argument to the quantifier is restricted by the first.\footnote{We are leaving out a lot of detail here which are not directly relevant to our discussion. We refer to the original paper.} Against this account, \citeA{knowltonDeterminersAreConservative2020} provides empirical psycholinguistic evidence showing that upon hearing a sentence `Every A is B', participants mentally group the A-entities, but not the B-entities or the A-and-B-entities as would be predicted by interface filtering:
\begin{claim}\label{claim:mentalgroup}
    Humans mentally group the entities in the restrictor argument but not the other arguments.
\end{claim}

The third type of explanation is \textit{ordered predication}.\footnote{This is related to the notion of \textit{restricted quantification} discussed in \citeA{knowltonPsycholinguisticEvidenceRestricted2023} (see also \citeA{ludlowLanguageFormLogic2022,lasersohnCommonNounsModally2021,petersQuantifiersLanguageLogic2006} for related proposals). The relation between these notions is not crucial for our purposes so we simply refer to the original papers.} This says that determiners do not in the first place express a relation between two sets. Rather, the role of the internal argument is to restrict the domain. This account can explain Claim \ref{claim:mentalgroup}. However, while the restricted quantification hypothesis provides a satisfactory semantics-internal analysis, it moves the evolutionary question upstream: why did determiners specialize to express restricted quantifiers?

Another kind of explanation argues that conservative quantifiers emerge because they are easier to learn. Under the assumption that learning encodes a bias for simpler representation, \citeA{vandepolQuantifiersSatisfyingSemantic2023} study whether conservative quantifiers are simpler according to an approximation of Kolmogorov complexity and a logical encoding inspired by the \textit{probabilistic Language of Thought} (pLoT) framework \cite{Quilty-Dunn_Porot_Mandelbaum_2023,piantadosiLogicalPrimitivesThought2016,piantadosiFourProblemsSolved2016,piantadosiBootstrappingLanguageThought2012,fodorLanguageThought1975}. Their findings support the following claim:
\footnote{Related universals have been investigated for artificial neural networks by \citeA{steinert-threlkeldEaseLearningExplains2020}. However, the setup in this paper does not directly test for conservativity.}
\begin{claim}\label{claim:LOT}
    Individual conservative quantifiers are on average simpler to encode in the logical language of \citeA{vandepolQuantifiersSatisfyingSemantic2023}.
\end{claim}
\noindent An implication of this explanation is that conceptual complexity is enough to prevent lexicalization.   

The picture of conservativity is further complicated by experimental work. \citeA{hunterConservativityLearnabilityDeterminers2013} show that children find a conservative quantifier easier to learn than a non-conservative one. However, the effect failed to replicate in \citeA{spenaderAreConservativeQuantifiers2019}. More recently,  
\citeA{knowltonNewEvidenceUnlearnability2022} and \citeA{knowltonStrengthConservativityEvidence2025} present a variety of empirical evidence, supporting the following claim:
\begin{claim}\label{claim:unable}
    English speakers are unable to acquire a new English determiner expressing a non-conservative quantifier.
\end{claim}
\noindent This is particularly surprising because, as already mentioned above,
\begin{claim}\label{claim:able}
    English speakers are able to learn new English verbs expressing non-conservative-like concepts, e.g., `outnumber'.
\end{claim}
\noindent The point of this claim is not that `outnumber' directly expresses non-conservative quantification, but rather that the concept it expresses seems prima facie as complex as a corresponding non-conservative quantifier, and therefore that explanations based solely on conceptual complexity would need to explain why `outnumber' can be learned easily. In sum, Claim \ref{claim:LOT} goes some of the way towards an explanation of conservativity, but is prima facie in tension with Claims \ref{claim:only} and \ref{claim:able}, because conceptual complexity as such is independent of semantic type, making the same predictions for determiners and verbs.

In sum, previous literature provides a complex picture of conservativity and related empirical phenomena. On the one hand, determiners expressing non-conservative quantifiers are arguably typologically unattested (Universal \ref{claim:cons2}) and very difficult to learn (Claim \ref{claim:unable}). On the other hand, non-conservative concepts can be learned more easily (Claims \ref{claim:able}). A previous model suggests that conservative concepts are simpler (Claim \ref{claim:LOT}), but no model explains the apparent sensitivity to grammar. The relation between the evolutionary accounts and the psycholinguistic evidence is unclear (Claim \ref{claim:mentalgroup}), and the best account of this evidence, the restricted quantification account, leaves the evolutionary question open. Finally, the restricted quantification account of conservativity moves the evolutionary question upstream but does not resolve it. In the next Section, we aim to develop an evolutionary account that resolves these apparent tensions.\footnote{\citeA{lucking_referential_2022} propose an alternative explanation of conservativity based on the idea that quantification happens within the noun phrase. This approach represents a more radical departure from the standard analyses, since it involves modifying the semantic of NPs, which makes it difficult to compare to other accounts. We leave a more systematic comparison to future work.}

\subsection{Applying the framework to quantificational meaning}\label{sec:tradeoffmodel}

In the previous section, we briefly presented previous explanations for conservativity and the challenges they face. We now apply our framework to develop a novel evolutionary account of conservativity. We propose that conservativity emerges as an adaptation of a compositional language containing determiners to the pressures for conceptual simplicity and communicative accuracy, when signals are structured and communication is pragmatic. More technical and implementation details on this Section are given in Appendix~\ref{appendix:conservativity}.

As described in Section~\ref{sec:model}, the semantics of the language is specified up to the meaning of the three quantifiers ($Q_1$, $Q_2$, and $Q_3$) and the component $\mathbb{C}_Q$ of the composition function that determines how quantifiers compose with their arguments. A possible language in this model is therefore obtained by specifying a value for these components. The components are specified as productions from probabilistic context-free grammars (PCFG), one for the lexicon and one for the composition function. The selection of conceptual primitives is a subset of the repertoire of \citeA{vandepolQuantifiersSatisfyingSemantic2023}. In terms of set construction, this includes the two basic arguments to the quantifiers, $L$ and $R$, the universe $\mathbb{U}$, and operators $\cup$, $\cap$, and $\setminus$ to construct other sets.

The grammar to define the meaning of individual quantifiers includes set-theoretic operations, Boolean operations, operators to deal with set sizes and comparisons between them, as well as the two arguments to the quantifier, $L$ and $R$. Although this is a simple grammar, it can encode a variety of quantifiers, such as `all' ($ | L \setminus R | = 0 $), `some' ($|L \cap R| > 0$), `no' ($ | L \cap R | = 0 $), `most' ($| L \cap R | > | L \setminus R | $), `only' ($| R \setminus L | = 0$), `exactly 1' ($| L \cap R| = 1$), `three or more' ($| L \cap R | > 1+1$), `not all' ($\neg \text{all}$), `some and not all' ($\text{some} \land \neg \text{all}$), `equi' ($|R| = |L|$), and others. Moreover, the inclusion of the Universe $\mathbb{U}^c$, which matches every object in $c$, makes it easy to construct a quantifier that depends on elements outside of $L \cup R$. Nonetheless, the conceptual grammar for encoding quantifiers' meanings poses some significant restrictions. For instance, since the grammar does not contain resources to refer to specific entities, we assume that meanings satisfy the universal of \textit{quantity} \cite{keenanSemanticCharacterizationNatural1986}.

The other unspecified component of the language is the part of the composition function that composes quantifiers with their arguments, $\quantcomp$. $\quantcomp$ composes a quantifier $Q$ with its two syntactic arguments $L$ and $R$. The grammar to define this part of the composition function includes set theoretic operations, boolean operations, and the quantifier argument $Q$. $\quantcomp$ also takes $L$ and $R$, the left and right syntactic argument of the quantifier. The simplest example of $\quantcomp$ is $Q(L)(R)$, which simply corresponds to composition with function application. A more interesting example is $Q(L)(R) \land Q(R)(L)$, which asserts that the quantifier is true for both orders of the two syntactic arguments. A third example is $Q(L \cup R)(L \cup R)$, which asserts that the quantifier is true when the union of the arguments is used for both the quantifier's arguments.\footnote{In principle, we could always explicitly prepend the definition of the body of $\quantcomp$ with $\lambda Q. \lambda L. \lambda R.$, but we omit this to avoid notational clutter. In order to make the search more efficient, we put a constraint that $L$ or $R$ have to appear in the body of $\quantcomp$. However, we do not put this constraint on the meaning of the three quantifiers. As we will see below having some trivial quantifiers is an interesting separate case.}

The main role of the composition function in this model is to introduce the possibility to share abstractions across quantifiers by passing them as arguments to the quantifier during composition. This can improve the expressiveness of the quantificational system while keeping the individual quantifiers simple. As an example of the interaction between the composition operator and the individual quantifiers, consider the following system:
{
\small
\begin{align}
\quantcomp &= 
Q ( \mathbb{U} ) ( L \cap R ) \label{eq:cexample} \\
Q_1 &= 
| L | = | R | \qquad
Q_2 = 
| L | > | R | \qquad
Q_3 = 
| R | = 1 \label{eq:endcexample}
\end{align}
}%
\noindent Quantifier $Q_1$ simply says that the two arguments have equal sizes. However, instead of composing the quantifier directly with its two syntactic arguments, the composition function uses the universe as the left argument and the intersection of the two syntactic arguments as the right argument (Eq.\ref{eq:cexample}). Therefore, after composition $Q_1$ effectively means that everything has both the properties expressed by $L$ and by $R$. We can obtain the effective meaning of a quantifier in the system by applying it to the composition function:
{\small
\begin{align}
    \quantcomp(Q_1) 
    &= 
    \left( |\mathbb{U}| = |L \cap R| \right)\\
    \quantcomp(Q_2) 
    &= 
    \left(| \mathbb{U} | > | L \cap R | \right) \\
    \quantcomp(Q_3) 
    &= 
    \left(| L \cap R | = 1 \right)
\end{align}
}%
From the point of view of the syntax-semantic interface, this also makes it possible to semantically specialize the syntactic arguments of the quantifiers. For instance, for $\quantcomp(Q)=(\mathbb{U}^c \setminus L)(R\setminus L)$, the left syntactic argument plays the role of a `negative' argument, i.e., its content is ignored. 

The implementation also makes a simplifying assumption about processing. Specifically, we assume that in the composition function arguments are applied to the quantifier argument progressively as they are given a value. In practice, this means that the first argument to $Q$ in the composition function cannot use $R$ (since it has not been given a value yet), but the second can use $L$ or $R$.
This assumption restricts the availability of $R$ to the second argument of $Q$, but not how it is used. Therefore, it does not force conservativity. For example, the following system is not conservative:
\begin{align}
    \quantcomp &= Q(L)(R\setminus L)\\
    Q_1 &= |R| = |L| \\
    \quantcomp(Q_1) &= |R\setminus L| = |L|
\end{align}
We call this an assumption of \textit{gradualness}, and come back to its consequences below.

In the following, we call the meaning expressed by the quantifier before composition ($Q_i$) its \textit{bare} meaning, and the meaning expressed by the quantifier after composition ($\quantcomp(Q_i)$) its \textit{effective} meaning. For similar reasons, we call the arguments passed to the effective meaning of the quantifier its \textit{unprocessed} arguments (the $L$ and $R$ from Table \ref{tab:composition}), and the arguments passed to the bare meaning the \textit{processed} arguments ($L$ and $R$ from Table \ref{tab:LoT} in the Appendix).

\subsection{Results}\label{sec:results}

\begin{figure}[t]
\centering
\begin{subfigure}[t]{0.49\textwidth}
    \centering
    \scalebox{0.8}{\includegraphics{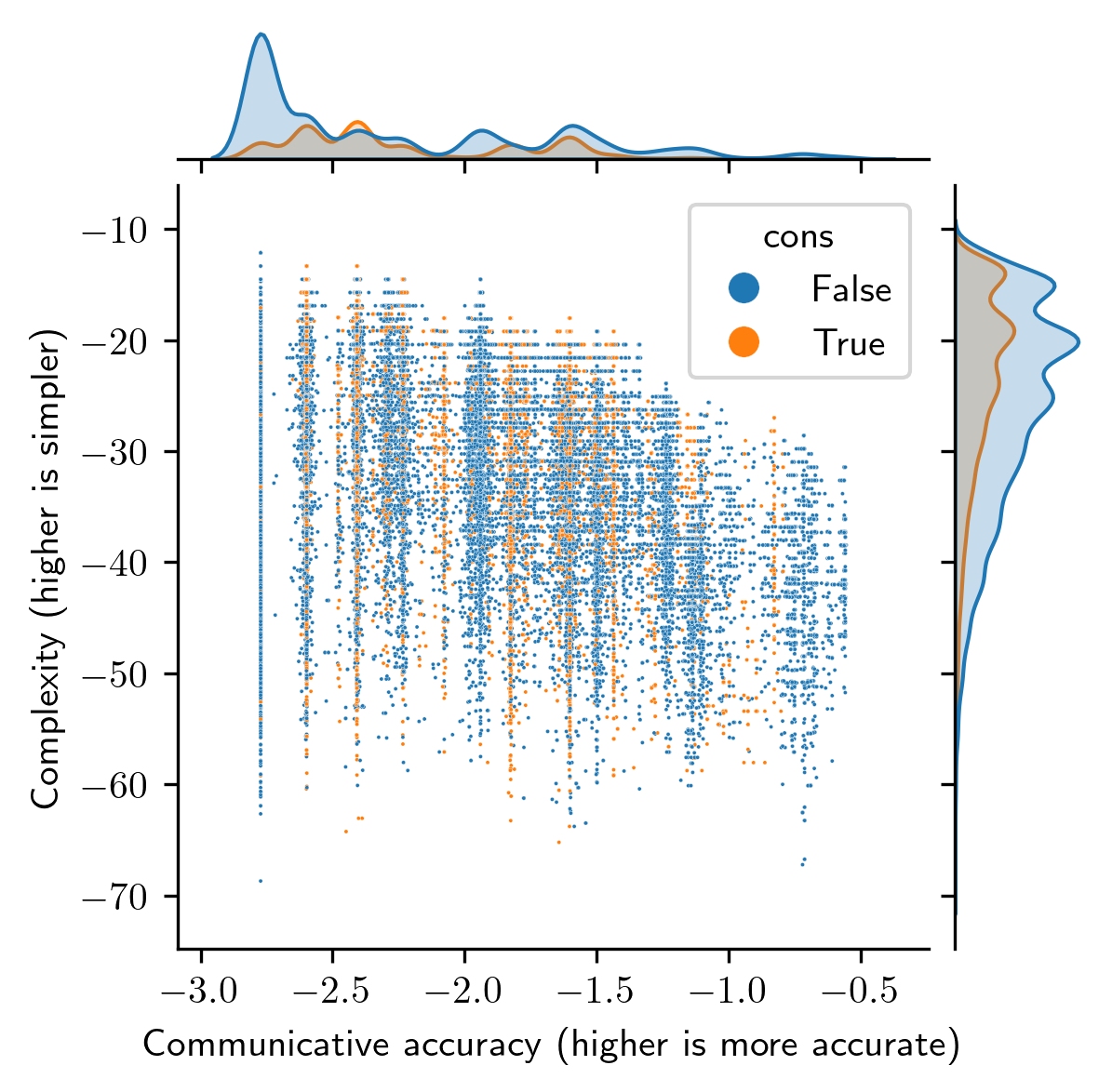}}
    \caption{Literal speakers.}
    \label{fig:tradeoff-a}
\end{subfigure}
\hfill
\begin{subfigure}[t]{0.49\textwidth}
    \centering
    \scalebox{0.8}{\includegraphics{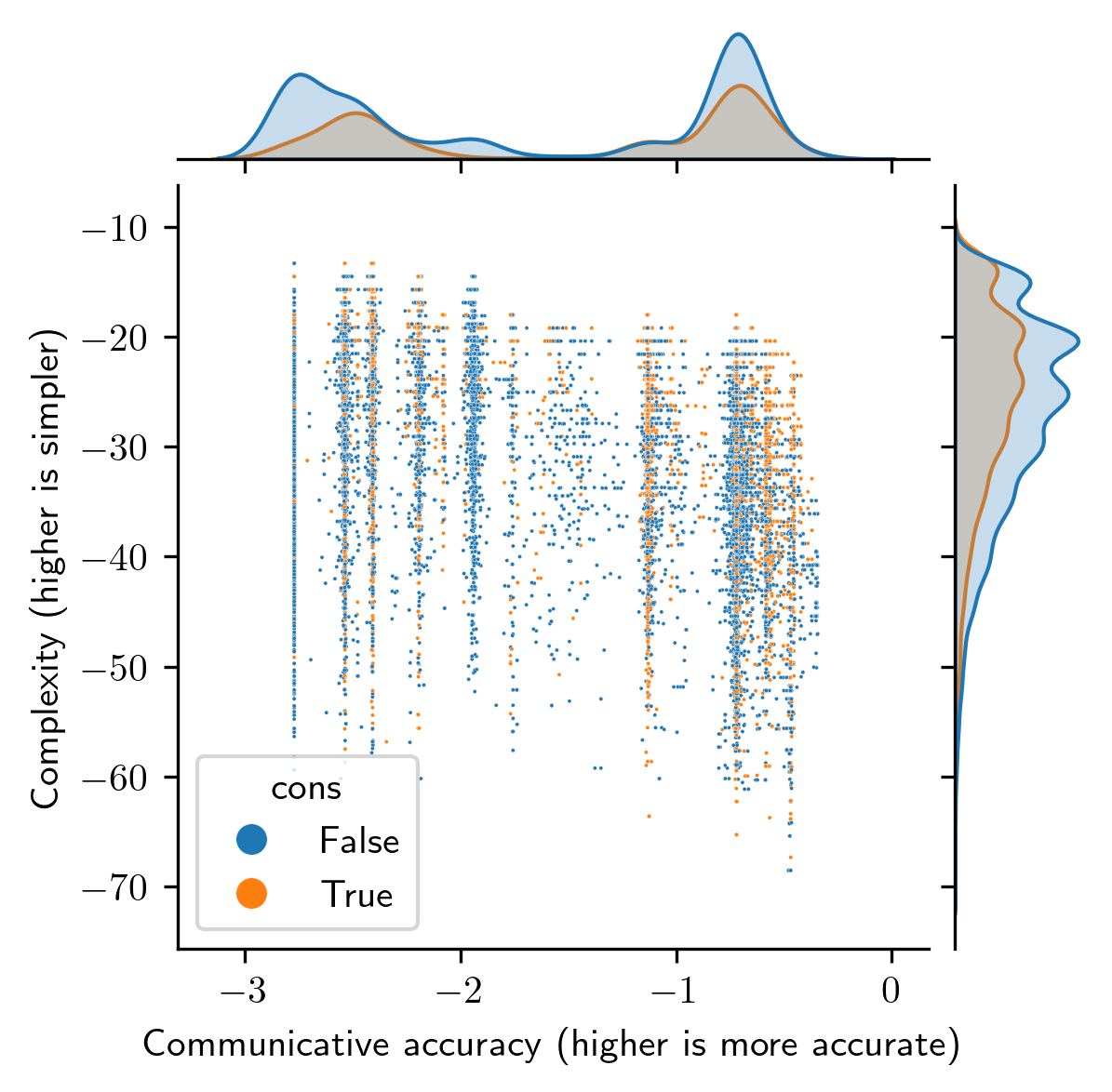}}
    \caption{Pragmatic speakers.}
    \label{fig:tradeoff-b}
\end{subfigure}
\caption{Languages sampled in the simulation, color coded by conservativity. Degenerate systems are excluded.  Details on the calculation of communicative accuracy are given in Appendix~\ref{appendix:technicalDetail}, and on complexity in \ref{appendix:conservativity}.}
\label{fig:resultsfull}
\end{figure}

In order to explore the space of possible languages, we use the C++ library Fleet \cite{yangOneModelLearning2022},
\footnote{We thank Steven Piantadosi for help with debugging, the suggestion to use Fleet for tradeoff analysis, and helpful discussions.}
to produce samples of languages close to the Pareto frontier of complexity and communicative accuracy.\footnote{All code for the simulations and analyses reported in this paper is available at 
[\textbf{Code will be made available upon acceptance}]} 
In line with previous work \cite{kempKinshipCategoriesLanguages2012}, we look at the Pareto frontier of languages with complexity and communicative accuracy as the dimensions of optimization. Since we are interested in Universal \ref{claim:globalcons}, we implement a check for conservativity at the level of the whole system of quantifiers. We are looking for languages where the truth of the effective meanings of all quantifiers only depends on the content of a \textit{restrictor}, which for each language can be either the denotation of the left unprocessed argument ($L$), or the denotation of the right unprocessed argument ($R$). See Appendix~\ref{appendix:setdependence} for more detail on this measure. Figure \ref{fig:resultsfull} shows the posterior samples for the literal and pragmatic speakers, color-coded by conservativity, on dimensions of conceptual complexity and communicative accuracy. More details on the proportion of conservative languages are given in Appendix~\ref{appendix:prior}.

\begin{figure}[t!]
    \centering
    \includegraphics[width=\textwidth]{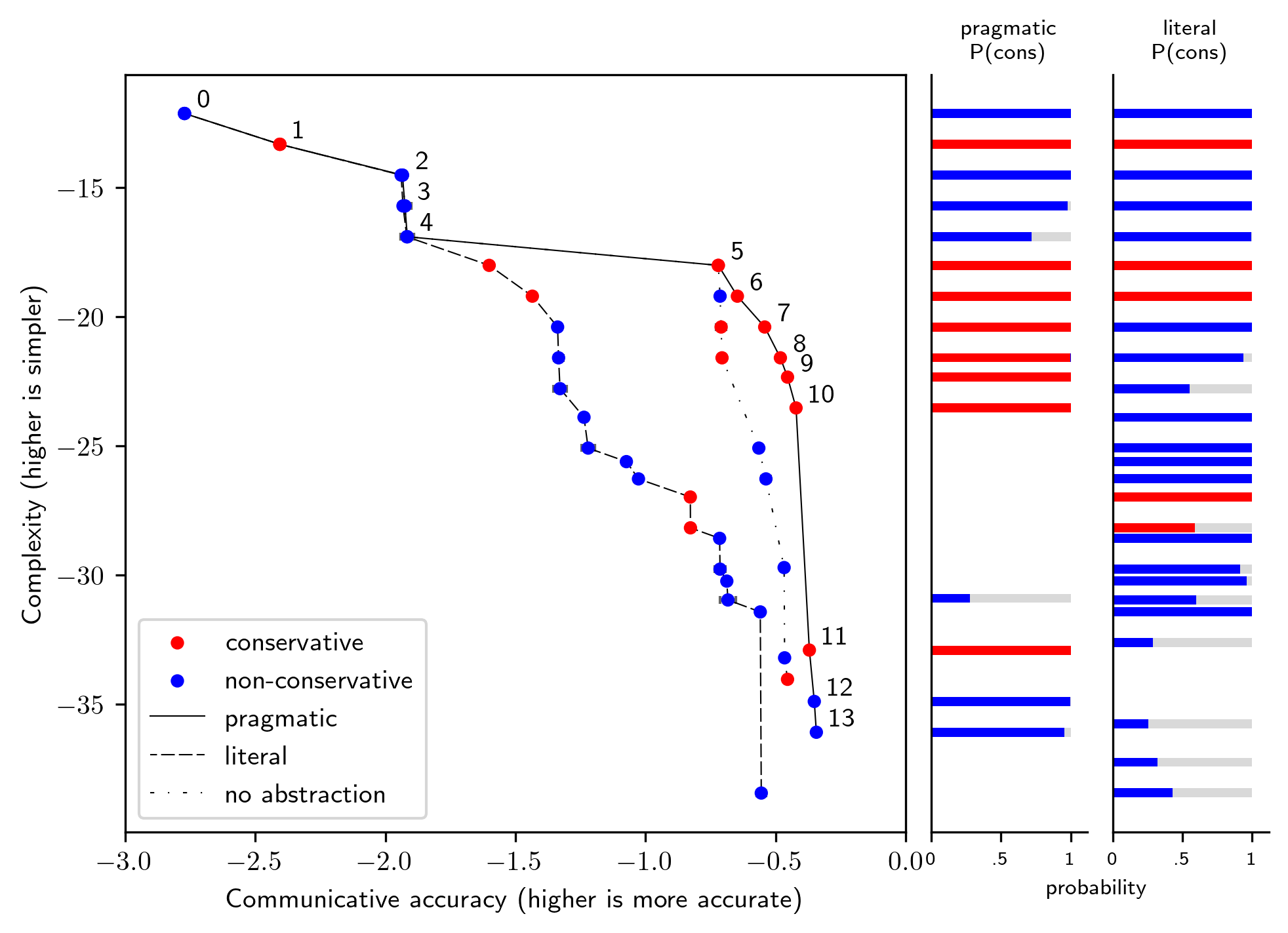}
    \caption{This Figure shows languages at the Pareto frontier of descriptive complexity and communicative accuracy for different pragmatic agents with a flexible composition function, literal agents with a flexible composition function, and pragmatic agents with composition fixed to function application. Details of these options are given in the main text. The side panels show uncertainty in the Pareto languages for each prior level. For each system, the SE of communicative accuracy was estimated from independent runs. Accuracies were then perturbed 2,000 times by Gaussian noise with these SEs, and the Pareto front was recomputed for each perturbation. For each prior level, the bars show the proportion of perturbations in which the language is conservative (red), non-conservative (blue), or there was no language (gray). This shows that the results discussed in the main body are robust to plausible perturbations. Error bars on the frontier points give 95\% intervals for accuracy, but most are smaller than the markers.}
    \label{fig:paretopragmatic}
\end{figure}

Figure \ref{fig:paretopragmatic} shows the languages at the Pareto front under various parameter settings. The full systems for each numbered language are reported in Appendix~\ref{appendix:results}. The most important result concerns the Pareto frontier with pragmatic agents and allowing arbitrary abstractions. Within this front, we can distinguish three groups, separated by large chasms of communicative accuracy (language 4 $\rightarrow$ 5) and conceptual complexity (language 10 $\rightarrow$ 11). 

The first group includes languages [0-4]. These are conceptually very simple, containing one or more `degenerate' quantifiers. They do not make use of any system-level abstraction, and/or they only manipulate the cardinality of bare arguments, $|L|$ and $|R|$. As a consequence of their extreme simplicity, they have low communicative accuracy. Due to this, they are not plausible candidates for realization in a natural language.

The second group includes languages [5-10] that strike a balance between complexity and communicative accuracy. They perform much better than languages in the first group for communication (note the x-axis gap between languages 4 and 5), and only at a small additional conceptual complexity cost. They can do this by introducing an abstraction at the level of their whole system of quantifiers which encodes conservativity. This abstraction corresponds to a semi-categorematic system of the kind introduced in Section~\ref{sec:categorematic}, and as a consequence all these languages are \textit{globally} conservative.

There is an interpretable reason for the advantage of conservative quantifiers: It is a simple and powerful way to generate entailment patterns between signals that can be exploited by the pragmatic process. In the simplest case, a quantified signal entails a non quantified one. For example, language 5 contains a single substantial quantifier encoding `all'. In this language, for all NPs $P$ such that $P(\text{1e})$, `all $P$s are target' entails non quantified signals such as `e1 is a target'. As a consequence, `e1 is a target' pragmatically conveys that `not all $P$s are targets'. For instance, in context \ref{eq:cexample} above, `$\text{target}(\text{1e})$' pragmatically conveys that not all negative numbers are target and not all even numbers are targets.

For systems with a single quantifier, abstracting conservativity into the composition function provides little advantage. However, with multiple quantifiers conservativity allows more complex interactions between quantified sentences that increase communicative accuracy further. Consider for instance language 8, which has the following effective quantifier meanings (encoded more efficiently thanks to the system-level abstraction $|R \cap L|$):
\begin{align}
    \quantcomp(Q_1) &= L \subseteq R & \text{All} \\
    \quantcomp(Q_2) &= (R \cap L = \emptyset) & \text{None} \\
    \quantcomp(Q_3) &= (|R \cap L| \geq 2) & \text{At least 2}
\end{align}
In this language, $Q_1(L)(R)$ asymmetrically entails $Q_3(L)(R)$ whenever $|L|\geq2$, and therefore if the extension of $L$ is in the common ground the sender can enrich $Q_3(L)(R)$ to $Q_3(L)(R) \land \neg Q_1(L)(R)$. In this case, abstracting conservativity in the composition operator reduces the global complexity without decreasing communicative accuracy.

The third and last group includes languages [11-13]. These languages are conceptually much more complex than the ones in group 2, but provide only very small additional gains in communicative accuracy. They tend to have more exotic quantifiers, e.g., $(|R \cap L|=0 \land |\mathbb{U}|> 2|L| )$ in language 12. Interestingly, these languages also encode conservativity in their composition function (either $Q(L, L \setminus R)$ or $Q(L, L \cap R)$), and their non-conservativity is rather a consequence of the fact that their quantifiers use the size of the universe set $\mathbb{U}$. It is plausible that the communicative utility of these quantifiers would decrease when the universe is very large compared to the denotation of the nouns in the language.

Figure \ref{fig:paretopragmatic} also shows that pragmatic enrichment is necessary for conservativity to be a successful strategy. At each level of language complexity, literal agents have much lower communicative accuracy than pragmatic agents. Notably, for literal agents conservativity does not emerge in a systematic way, but rather the Pareto frontier has a mix of conservative and non-conservative systems. This Figure also confirms that the large gap in accuracy between the first and second groups of the pragmatic Pareto frontier is due to the fact that conservativity exploits pragmatic communication. The accuracy gap between literal and pragmatic is again much smaller for the languages in the third group (complexity greater than $\approx -30$).


Finally, Figure \ref{fig:paretopragmatic} shows that the ability to form abstraction is also required for the evolution of conservativity. At each level of complexity in group 2, languages without any abstraction in the composition function (more specifically, languages where $\quantcomp$ only consists of a $Q$ composing with bare $R$ and/or $L$) perform communicatively worse. With this constraint, the languages on the Pareto frontier are mostly non-conservative. In conclusion, conservativity is the \textit{simplest} abstraction that exploits the possibility of compressing the system of determiners to improve communication.    


\subsection{Discussion}\label{sec:discussion}

Previous empirical and theoretical research posed several constraints on a satisfactory account of conservativity. We have seen that restricted quantification accounts for many of the empirical claims \cite{knowltonPsycholinguisticEvidenceRestricted2023}, but does not provide an evolutionary explanation. On the other hand, evolutionary accounts struggled to account for the empirical data. We claim that the account we presented above fares better. 

First, our account solves the apparent tension between the experimental results showing the unlearnability of non-conservative quantifiers (Claim \ref{claim:unable}) and the pLoT models that predict non-conservative quantifiers to be generally learnable \cite{vandepolQuantifiersSatisfyingSemantic2023}. If English quantifiers are conservative because the compositional system only gives them access to certain abstractions, it will be much harder to learn a new English quantifier with a non-conservative meaning, even when this quantifier is conceptually not very complex. 

Second, our account alleviates the tension between the absence of non-conservative quantifiers (Universal \ref{claim:globalcons}) and the presence of non-conservative-like verbs (Claims \ref{claim:able} and \ref{claim:only}), which has been hard to reconcile for previous evolutionary accounts. In our account, conservativity is a consequence of a system-level compression that happens at the level of the \textit{system} of quantifiers. Once $\quantcomp$ provides arguments that makes quantifiers conservative, adding a non-conservative quantifier to the system becomes difficult. Therefore, it is not conceptual complexity of single words that prevents the lexicalization of conservativity, but rather system-level complexity. The same constraint does not (necessarily) evolve in the verbal domain, dissociating the two facts. A full explanation would also need to explain why a similar system-level abstraction encoding conservativity does not evolve in the verbal domain. The natural answer is that abstractions in the composition function conditioned on types, such as we claim conservativity to be, are applied uniformly to all arguments. However, most verbs do not express concepts involving relations between sets, and therefore conservativity cannot be stated in general for verbs.

Third, our model alone does not as such explain Claim \ref{claim:mentalgroup}. However, it also is not in tension with it. The two-level organization proposed in this paper complicates the linking function from the theory to the psycholinguistic predictions. An assumption that would link our model with Claim \ref{claim:mentalgroup} is that participants mentally group the sets constructed by the composition function, or the ones constructed by the individual quantifiers, or both.


One modeling assumption we made was that arguments are only processed when they have been given a specific value. This assumption can naturally be motivated as a processing constraint, i.e., the quantifier must process its arguments progressively and only arguments that have been given a value can be processed. 
However, it does not substantially change the results. Lifting this assumption only adds one Pareto language. This is a non-conservative language that is significantly more complex than language 10 for only a small increase in accuracy. It still lies before the large jump in complexity of the third group. These are the effective meanings of the quantifiers in this language:
\begin{align}
    \quantcomp(Q_1) &= (|R \cap L| > 1) & \text{At least 2} \\
    \quantcomp(Q_2) &= (|R| = |L|) & \text{Equinumerous} \\
    \quantcomp(Q_3) &= (|R \cap L| = 1) & \text{Exactly 1}
\end{align}
This system is conservative except for the quantifier expressing the equinumerosity of two sets. 

Despite its advantages, the model we considered above is limited in various ways by computational feasibility. The code base was designed to be flexible enough to accommodate other semantic domains, at the expense of some efficiency. As a consequence, combinatorial explosion makes it unfeasible to evolve more than three quantifiers at once. While the advantage of conservativity is already clear with three quantifiers, it is likely that with more quantifiers the conservativity abstraction becomes even more useful to compress a system compared to a system without any abstractions. Moreover, the usefulness of additional quantifiers depends on the size of the domain. With a context size of 5, there may not be enough distinctions in possible set sizes to justify a large inventory of quantifiers. Therefore, we leave an analysis of the implications of a larger available repertoire of quantifiers to future work.

The reader might also wonder whether the choice of lexicon and the conceptual primitives for the LoT matters for the results (see Table \ref{tab:language} in Appendix~\ref{appendix:technicalDetail}). The selection of lexicon and primitives involved some arbitrary choices. A systematic exploration of the effects of each choice would be computationally unfeasible. However, it is worth noting as discussed above in Section~\ref{sec:results} that only a minority of the systems sampled in the simulation are conservative. Moreover, Figure~\ref{fig:resultsfull} shows that the predominance of non-conservative system holds across degrees of complexity, further suggesting that the specific set of primitives in the simulation does not advantage conservativity in a straightforward way. Finally, the chosen inventory of conceptual primitives are a standard basis that provides a natural default choice. Therefore, we leave a systematic investigation of the choice of primitives to future work.

\section{Conclusions}\label{sec:conclusions}

We have argued that the interaction of lexical meanings and the composition operator can play an essential role in explaining semantic universals, and therefore that the full picture offered by formal semantics should be considered when studying language evolution. We have presented a modeling framework that implements this perspective, and shown that it can account for an important semantic universal better than previous accounts.

The model presented in this paper is static, i.e., it does not explicitly model the evolutionary mechanism that would lead to the development of languages at the Pareto front. In particular, we have assumed that while being Pareto optimal, languages with extremely low communicative accuracy would in practice never emerge (Group 1 in Figure~\ref{fig:paretopragmatic}). Pareto front analysis is a standard methodology in the study of semantic universals \cite{uegakiInformativenessComplexityTradeoff2024,kempSemanticTypologyEfficient2018,zaslavskySemanticCategoriesArtifacts2019,denicComplexityInformativenessTradeoff2021}. \citeA{carrSimplicityInformativenessSemantic2020} systematically studies the relation between these two levels of analysis. Future work can extend these simulations with an explicit evolutionary model.

The overall goal of this paper was to demonstrate the usefulness of a deeper integration of computational language evolution research and formal semantics. The modeling framework we developed could be applied to other semantic universals. Potential candidates include any universal that involves semantic specialization of syntactic arguments or composition between specific combinations of types. Moreover, while the model presented in this paper is not as such a model of the emergence of compositionality, a similar model could target the evolution of the attested modes of composition. The intensional formulation of all lexical meanings in the model could also be used to study universals in the domain of modality. Finally, by adapting the structure of contexts a similar model could be used to study the way sentences express event structure. 

There are reasons to take compositional structure seriously in evolutionary accounts beyond the one we discussed in this paper. First, the contribution of a given lexical entry to communicative accuracy can only be evaluated in the context of the whole language, especially when pragmatic inferences come into play that enrich a signal depending on other signals in the language, as happens in the case study above. Therefore, it is important to relate the isolated meaning of a lexical entry to its contribution across the utterances in the language, which requires composition. Second, the modes of composition that have been identified in natural language, e.g., saturation, function application, type shifting, etc., can only manipulate appropriate lexical meanings, e.g., higher-order functions. Therefore, a model of natural language compositionality must include an appropriate lexical semantics. Third, the evolution of some universals involves the co-evolution of different components of language semantics, e.g., the co-evolution of the composition function and the meanings of a set of lexical entries to encode semantic specialization of a syntactic argument. Finally, the most practical reason for considering lexicon and composition jointly is that, as we have argued for conservativity, universals of lexical semantics might be better characterized semi-categorematically in terms of the composition operation.

The results of this kind of modeling can also feed back into formal semantics. For instance, standard semantic analyses are usually categorematic, restricting the composition function to few generic modes of composition. Less `logical' constraints like conservativity are either specified case-by-case in the lexical entries, or simply stated as generalizations across a grammatical category, e.g., restricted quantification \cite{knowltonPsycholinguisticEvidenceRestricted2023}. This is in large part motivated by the worry of making semantic analysis vacuous: with arbitrary and ad-hoc modes of composition, any meaning can be derived. This worry is justified if there is no principled method to give an overall simpler analysis by balancing the complexity of the lexical meanings and the composition function. However, if the composition function describes how we construct the meaning of a constituent based on its sub-constituents, semantic specialization of specific arguments (e.g., `The internal argument of a quantifier is a restrictor') belongs to it. We have shown above how evolutionary modeling can provide support for specific ways of distributing semantic work between composition and the lexicon.

\section*{Acknowledgments}

Many thanks to Steven Piantadosi, Tyler Knowlton, Giorgio Sbardolini, Milica Denic, Heming Strømholt Bremnes, Jakub Szymanik, Michael Franke, Polina Tsvilodub, Nadja Gruhler, Marieke Woensdregt, Bob van Tiel, and Terry Regier for useful discussions. We also thank Steven Piantadosi for help with debugging, and the suggestion to use Fleet for tradeoff analysis.

\bibliographystyle{apacite}
\bibliography{bibliography} 

\newpage
\appendix

\section{Technical detail on model setup}\label{appendix:technicalDetail}

\subsection{World and communication}\label{subappendix:worldandcomm}

Each object consists of an integer $v$ between -10 and 10, representing a feature that identifies the object,\footnote{The structure of this first component might be different depending on the semantic domain under study.} and a Boolean $t$, which specifies whether the object is a target or a distractor:
\begin{equation}
     c = \left\langle \tup{ v_1, t_1 }, ..., \tup{v_5, t_5} \right\rangle
\end{equation}
Call $\prod_{v}c$ the tuple $\left\langle v_1, ..., v_5\right\rangle$ and similarly for $\prod_{t}c$. $c$ is constructed such that $\prod_{v}c$ is always increasing. $\prod_{v}c$ is in the common ground, i.e., observed both by $S$ and $R$, while $\prod_{t}c$ is only observed by $S$. For instance, this is one possible context:
\begin{equation}
    \left\langle
        \tup{-8, \operatorname{True}}, 
        \tup{-4, \operatorname{False}}, 
        \tup{-1, \operatorname{False}}, 
        \tup{0, \operatorname{True}}, 
        \tup{6, \operatorname{True}} 
    \right\rangle
\end{equation}
Contexts are generated by uniformly sampling five elements (without replacement) from $\{-10, \ldots, 10 \}$ and five Boolean values, and collecting these values appropriately into tuples. Each common ground is compatible with $2^5=32$ possible assignments of values to $\prod_t c$ (the boolean components).

Communicative success is defined as the negative surprisal of the true context for $R$ after updating the common ground (the interpretation function and pragmatic enrichment are defined below):
\begin{equation}\label{eq:successmeasure}
    \mathbb{U}\infdivx{s}{c} = 
    \log(P(c \mid s)) = -\log\left(\left| \left\{ c' \mid \prod_{v} c' = \prod_{v} c \land \eval{s}{c'}_\text{prag} \right\} \right| \right)
\end{equation}

\subsection{Semantics}\label{subappendix:semantics}

The lexicon along with the type of each word is presented in Table \ref{tab:language}. If $\alpha : \tup{s, \tau}$ is a bare lexical entry, then $\eval{\alpha}{c} : \tau$ is its extension in context $c$, i.e., $\alpha(c)$. Composability is determined solely by the semantic types of the constituents.

\begin{table}[t]
    \centering
    {\footnotesize
    \begin{tabular}{l|l|l}
        Name & Meaning & Semantic type \\
        \midrule
        1e, ..., 5e & Proper names & $\tup{s,e}$ \\
        thing, even, prime, target, distractor & Unary predicates & $\tup{s,\tup{e,t}}$ \\
        greater than, equal & Binary predicates & $\tup{s,\tup{e,\tup{e,t}}}$ \\
        notP & Predicate negation & $\tup{s,\tup{\tup{e,t},\tup{e,t}}}$ \\
        andP, orP & Predicate modifiers & $\tup{s,\tup{\tup{e,t},\tup{\tup{e,t},\tup{e,t}}}}$ \\
        $Q_1$, $Q_2$, $Q_3$ & Quantifiers & $\tup{s,\tup{\tup{e,t}, \tup{\tup{e,t},t}}}$
    \end{tabular}
    }
    \caption{Lexicon of one language, along with linguistic category and semantic type. Meanings are defined in the natural way. Proper names refer to the indices of the objects in the context tuple. $e$ is a shorthand for the type of the elements of the context, i.e., $\left( \mathbb{N} \times t \right)$. All meanings are functions $\tup{s, \tau}$ that take a context and return an extension of type $\tau$ in that context.\label{tab:language}}
\end{table}

Interpretations are encoded in an intensional type system that is standard in formal semantics, which includes every type $\tup{s, \tau}$ with $\tau$ defined recursively as:
\begin{equation}
    \tau ::= e \mid t \mid \tup{\tau, \tau}
\end{equation}
where $\tup{s,\tau}$ is the type of functions from a context to an object of type $\tau$, $t$ is the type of truth values, and $e$ is the type of entities (of type $\mathbb{N} \times \operatorname{Bool}$). The system must be intensional because the truth of an expression has to be evaluated on each possible context. 

Because of the constraints imposed by the type system of C++, the composition function is in fact implemented as a function manipulating objects of a sum type containing any type at most as complex as the most complex lexical type. This assumes that the composition function will not arbitrarily grow types much beyond the ones represented in the lexicon. This is true in the current model but might be modified in future work. We leave out these complication for ease of exposition.

\subsection{Pragmatics}\label{subappendix:pragmatics}

The literal meaning of a signal $s$ in true context $c$ is enriched as follows. Construct the set $U$ of alternative signals $u$ such that restrict the common ground $\prod_v c$ more than $s$: 
\begin{equation}
\forall c' : \left( \prod_v c' = \prod_v c  \right) . \eval{u}{c'} \rightarrow \eval{s}{c'} \land \eval{s}{c'} \not\rightarrow \eval{u}{c'} 
\end{equation}
The interpreter assumes that the speaker produces the most informative possible signal. Therefore, if $c$ verified a signal in $U$, $s$ would never be produced. $\eval{s}{c}$ then conveys the negation of each of its strictly stronger alternatives:
\begin{equation}
    \eval{s}{c}_{\text{prag}} = \eval{s}{c} \land \bigwedge_{u \in U} \neg \eval{u}{c}
\end{equation}
Since $c$ is false for all alternatives, this conjunction is true at least of $c$. For example, assuming that $Q_1$ is assigned meaning `all', `target(1e)' (= 1e is a target) produces an implicature that `$Q_1$(thing)(target)' (= everything is a target) is false, because the latter unidirectionally entails the former.

While simple, this strategy can have complex effects on what signals convey. In contrast to e.g., lexical Horn scales \cite{gazdarPragmaticsImplicaturePresupposition1979}, each signal may implicate the negation of structurally different signals or signals with a different argument order. Moreover, entailments are calculated relative to the common ground. So for instance, `1e is a target' entails `some negative elements are targets' if $\prod_v c = \tup{-3, -1, 5, 7, 9}$, but not if $\prod_v c = \tup{1, 3, 5, 7, 9}$. 

Because each signal is contrasted with each other signal, one might wonder whether this strategy for pragmatic enrichment encounters the problem of symmetric alternatives. For instance, both $\exists \land \neg \forall$ and $\forall$ imply $\exists$, and therefore $\exists$ would convey both $\neg \left( \exists \land \neg \forall \right)$ and $\neg \forall$, which contradict each other. However, such a scenario could not arise in the simulation. Bare $\exists$ would never be produced if a more informative signal was available. This is a consequence of the decision to not model partial knowledge; the speaker always knows the full context. Future work can model partial knowledge and calculate implicatures based only on structural alternatives \cite{foxCharacterizationAlternatives2011,katzirStructurallydefinedAlternatives2007}, or integrate probabilistic models of pragmatics such as the Rational Speech Act model \cite{frankeProbabilisticPragmaticsWhy2016,frankeTheorydrivenStatisticalModeling2020,scontrasPracticalIntroductionRational2021}.


\subsection{Communicative accuracy}

Since agents deterministically pick the most informative signal in context, communicative accuracy is calculated as:
\begin{align}    
    \mathbb{U}(\lang) 
    &= \sum_{c} P(c) \max_{s \in \lang} \mathbb{U}\infdivx{s}{c}\label{eq:languageEU} \\
    &= \sum_c P(c) \log P\left(c \mid f_{\lang}(c)\right) \\
    &= \sum_c P(c) \log P(c) - \sum_c P(c) \log P(f_{\lang}(c))
\end{align}
where $f_{\lang}$ picks out for a context $c$ the signal in $\lang$ that maximizes utility. 
For pragmatic agents see \ref{subappendix:pragmatics}. Ties in signal selection are broken with the shortest signal. $\mathbb{U}(\lang)$ is then the negative conditional entropy of the context given the signal, and it measures (the negative of) the expected uncertainty left about the context after the signal is received. Since there are $P^{21}_5 \times 32=651,168$ possible contexts, and infinitely many signals in the language, we approximate $\mathbb{U}(\lang)$ by sampling 500 contexts and finding the average utility of the context-wise best signal (up to a fixed depth).

\section{Implementation of case study of conservativity}\label{appendix:conservativity}

The grammar for specifying the quantifiers is described in Table \ref{tab:LoT}. The space of possible values for $\quantcomp$ is defined by the CFG described in Table \ref{tab:composition}. These are based on the grammar in \citeA{vandepolQuantifiersSatisfyingSemantic2023}. The $\iota$ operator is excluded to make the exploration of hypothesis space faster. The individual items in the context can be picked out in communication by their lexicalized proper names. Complexity in this PCFG is the negative log probability of a derivation. Rule probabilities are the normalized weights within each return type. 

\begin{table}[t]
    \centering
    {\small
    \begin{tabular}{l|l|l|l}
        Name & Meaning & Semantic type & Weight \\
        \midrule
        $\mathbb{U}^c$ & Universe & $\tup{e, t}$ & 10 \\
        $ \cup  $, $ \cap  $, $ \setminus  $ & Predicate operations &  $\tup{\tup{e,t},\tup{\tup{e,t},\tup{e,t}}}$ & 1 \\
        $\neg \,  $ & Boolean negation & $\tup{ t, t }$ & 1 \\
        $ \land  $, $ \lor  $ & Boolean operations & $\tup{ t, \tup{t, t} }$ & 1 \\
        $L$, $R$ & Quantifier arguments & $\tup{ e, t }$ & 10 \\
        $|\cdot|^{c}$ & Cardinality & $\tup{ \tup{e, t}, \mathbb{N} }$ & 10 \\
        $0, 1$ & Numbers & $\mathbb{N}$ & 10 \\
        $ > ,  = $ & Numerical comparisons & $\tup{\mathbb{N}, \tup{\mathbb{N}, t }}$ & 10 \\
        $ + ,  - $ & Numerical operations & $\tup{\mathbb{N}, \tup{\mathbb{N}, \mathbb{N}}}$ & 1 \\
    \end{tabular}
    }
    \caption{Terminals of the CFG to construct the meaning of quantifiers. The CFG is determined by composability via function application. All operations are interpreted intensionally with the context as the first argument, but the dependence is not written below when not relevant.}
    \label{tab:LoT}
\end{table}

\begin{table}[t]
    \centering
    {\small
    \begin{tabular}{l|l|l|l}
        Name & Meaning & Semantic type & Weight \\
        \midrule
        $\mathbb{U}^c$ & Universe & $\tup{e, t}$ & 10 \\
        $ \cup  $, $ \cap  $, $ \setminus  $ & Predicate operations &  $\tup{\tup{e,t},\tup{\tup{e,t},\tup{e,t}}}$ & 1  \\
        $\neg \,  $ & Boolean negation & $\tup{ t, t }$ & 1 \\
        $ \land  $, $ \lor $ & Boolean operations & $\tup{ t, \tup{ t, t} }$ & 1 \\
        $L$, $R$ & Composition arguments & $\tup{ e, t }$ & 10 \\
        $Q$ & Quantifier argument & $\tup{\tup{e,t}, \tup{\tup{e,t},t}}$ & 10 \\
    \end{tabular}
    }
    \caption{Terminals of the grammar to construct the meanings of the component of the composition function that composes a quantifier with its arguments. The CFG is determined by composability via function application.\label{tab:composition}}
\end{table}

The composition function in this model presented in this paper distinguishes composition of a quantifier with its arguments, encoded by the function $\quantcomp$, with every other composition, which is done simply by (intensional) function application. Formally, for any types $\pi$ and $\mu$, if $\alpha$ is a branching node with daughters $\beta: \tup{s, \tup{\pi, \mu}}$ and $\gamma: \tup{s, \pi}$, $\eval{\alpha}{c}$ is defined as follows:
\begin{equation}\label{eq:compositionfunction}
\eval{\alpha}{c} = 
\begin{cases}
\quantcomp(\eval{\beta}{c})(\eval{\gamma}{c}) & \text{if } \vdash \beta: \tup{s,\tup{\tup{e,t}, \tup{\tup{e,t}, t}}} \\
\eval{\beta}{c} \left( \eval{\gamma}{c} \right) & \text{else.}
\end{cases}
\end{equation}

\section{Conservativity of prior samples}\label{appendix:prior}

\begin{figure}[t]
    \centering
    \includegraphics[width=\textwidth]{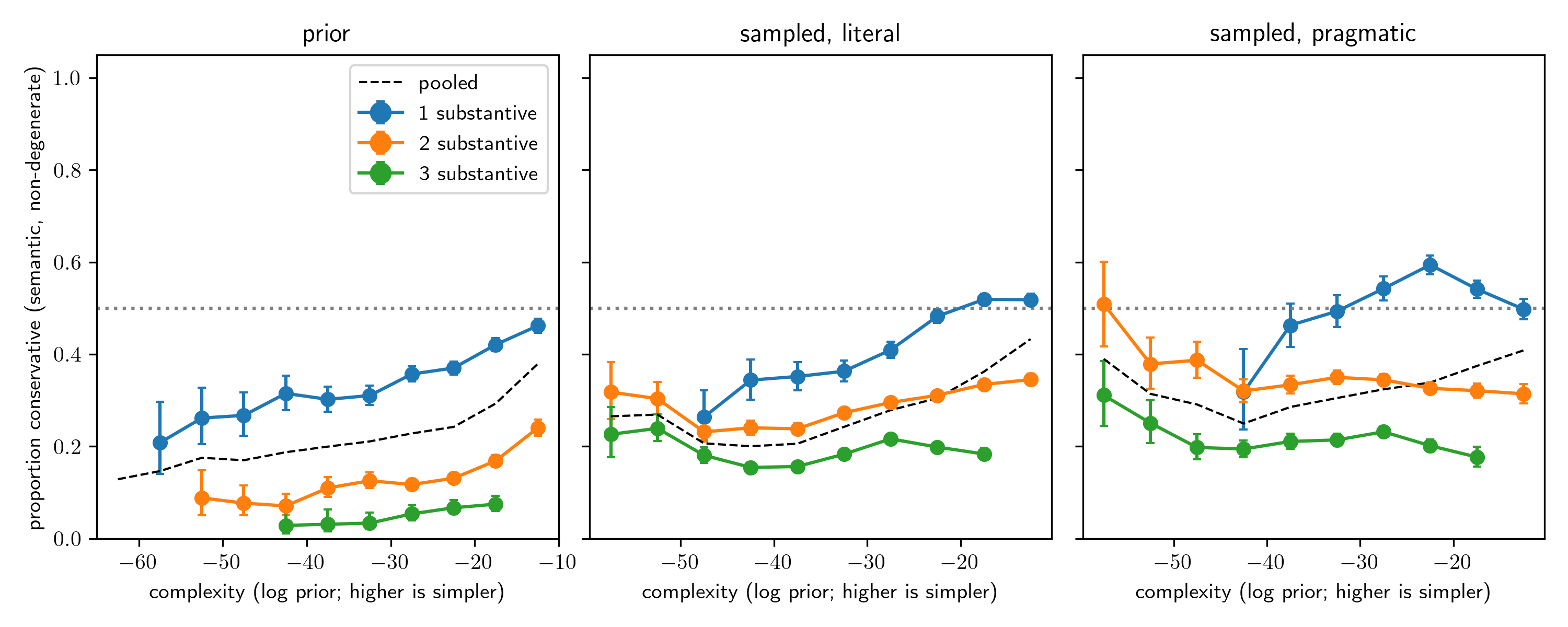}
    \caption{This Figure shows the proportion of conservative systems from 50'000 samples from the PCFG defining the prior over semantic systems, as well as for the simulation samples. Systems are split by the number of non-trivial quantifiers. Bins with fewer than 100 systems are omitted. The plot excludes degenerate systems. Error bars are 95\% Wilson score confidence intervals.}
    \label{fig:prior}
\end{figure}

Figure \ref{fig:prior} shows the frequency of conservativity in samples from the PCFG that produces the semantic systems and in the simulations presented in the main paper. The Figure shows that the choice of conceptual primitives does not provide a prior advantage to conservative systems, especially ones with 2 or 3 non-degenerate quantifiers. The plot also shows that the proportion of conservative quantifiers increases in the simulation samples. 

\section{Details on simulations and  results}\label{appendix:results}

We ran the simulation with the following parameter settings:
\begin{enumerate}
    \item $|c|=5$
    \item 500 simulated games to estimate the utility of each language.
    \item Maximum signal depth of 3.
    \item 200000 samples from the posterior for each run.
    \item Pragmatic or literal speakers $\times$ likelihood weights of 20, 30, and 40, for a total of 6 runs.
\end{enumerate}

After sampling, we also partition the sampled languages by semantic equivalence and take a mean within each cell. This reduces the variance in the estimation of the loglikelihood. In order to better construct the Pareto front, we also refine the estimates of communicative accuracy as follows. We first construct the Pareto front with the initial database. Then, we check if any of the points has fewer than 10 elements in its corresponding partition element. For the ones that do, we re-evaluate the communicative accuracy by simulating 5000 games and recalculate the Pareto front. We repeat this process iteratively until all points on the Pareto front are estimated based on at least 5000 sampled games. At the end, we remove languages with redundant operations, such as $+0$, which cannot be on the Pareto frontier and therefore had to be an approximation error.

We use weighted likelihood to explore different parts of the Pareto frontier.  This likelihood effectively encodes the mean proportional probability of the true context for the listener after receiving the signal. This is because:
$$
\exp\left( \frac{1}{|N|} \sum_{i=1}^{N} \log(P(c_i \mid f_{\lang}(c_i)) \right) \propto \prod_{i=1}^N P(c_i \mid f_{\lang}(c_i))
$$

\begin{table}[t]
\centering
\small
\begin{tabular}{r l l l l}
& & $Q_1$ & $Q_2$ & $Q_3$ \\
\toprule
 0&$Q\!\left(R,\, L\right) $ & $ \top $ & $ \top $ & $ \top $ \\
1&$Q\!\left(R,\, L\right) $ & $(1 = \lvert R \rvert^{ c })$ & $ \top $ & $ \top $ \\
2&$Q\!\left(R,\, L\right) $ & $ (\lvert R \rvert^{ c } = \lvert L \rvert^{ c }) $ & $ \top $ & $ \top $ \\
3&$Q\!\left(L,\, R\right) $ & $ (\lvert R \rvert^{ c } > \lvert L \rvert^{ c }) $ & $ (\lvert L \rvert^{ c } > 0) $ & $ \top $ \\
\midrule
4&$Q\!\left(L,\, R\right) $ & $ (\lvert L \rvert^{ c } > \lvert R \rvert^{ c }) $ & $ (\lvert L \rvert^{ c } > 1) $ & $ (1 = \lvert R \rvert^{ c }) $ \\
5&$Q\!\left((L \setminus R),\, L\right) $ & $ \top $ & $ \top $ & $ (\lvert L \rvert^{ c } = 0) $ \\
6&$Q\!\left(R,\, (L \setminus R)\right) $ & $ (0 = \lvert R \rvert^{ c }) $ & $ \top $ & $ (\lvert R \rvert^{ c } > 1) $ \\
7&$Q\!\left(L,\, (L \setminus R)\right) $ & $ (\lvert R \rvert^{ c } > 1) $ & $ (1 > \lvert R \rvert^{ c }) $ & $ (\lvert R \rvert^{ c } = 1) $ \\
8&$Q\!\left((R \cap L),\, L\right) $ & $ (\lvert L \rvert^{ c } = \lvert R \rvert^{ c }) $ & $ (\lvert L \rvert^{ c } = 0) $ & $ (\lvert L \rvert^{ c } > 1) $ \\
9&$Q\!\left((L \setminus R),\, L\right) $ & $ \neg (\lvert L \rvert^{ c } > 1) $ & $ (1 > \lvert L \rvert^{ c }) $ & $ \top $ \\
\midrule
10&$Q\!\left(R,\, (L \setminus R)\right) $ & $ (\lvert R \rvert^{ c } > 1) $ & $ (\lvert R \rvert^{ c } = 0) $ & $ \neg (\lvert R \rvert^{ c } > 1) $ \\
11&$Q\!\left(L,\, (L \setminus R)\right) $ & $ (1 > \lvert R \rvert^{ c }) $ & $ \neg (\lvert R \rvert^{ c } > 1) $ & $ (\lvert (L \cap (\mathbb{U}^{c} \setminus R)) \rvert^{ c } > 1) $ \\
12&$Q\!\left(L,\, (R \cap L)\right) $ & $ ((0 = \lvert R \rvert^{ c }) \land (\lvert \mathbb{U}^{c} \rvert^{ c } > (\lvert L \rvert^{ c } + \lvert L \rvert^{ c }))) $ & $ (\lvert L \rvert^{ c } = \lvert R \rvert^{ c }) $ & $ \top $ \\
13&$Q\!\left(L,\, (R \cap L)\right) $ & $ ((\lvert R \rvert^{ c } > 1) \land (\lvert \mathbb{U}^{c} \rvert^{ c } > (\lvert L \rvert^{ c } + \lvert L \rvert^{ c }))) $ & $ (\lvert L \rvert^{ c } = \lvert R \rvert^{ c }) $ & $ (0 = \lvert R \rvert^{ c }) $ \\
\bottomrule
\end{tabular}
\caption{All languages at the main Pareto frontier in Figure~3. 
Note that any composition function where one of the two arguments is a plain $R$ or $L$ can be re-written as a system satisfying the gradualness assumption without changing complexity and communicative accuracy. 
Moreover, for the systems where the first argument of $\mathcal{C}^c_Q$ is not used, it does not matter conceptually whether the second argument is $R \setminus L$ or $L \setminus R$, since in each signal involving quantifiers the syntactic arguments can be provided in the switched order. However, it matters in the simulation because of the depth bound: since the quantifier itself consumes one level of depth, right arguments of a quantifier can be 1-level deeper than left arguments.}
\end{table}

\section{Identification of conservative systems}\label{appendix:setdependence}

The larger number of systems in the simulation makes manual annotation impossible, therefore we compute conservativity automatically. Prima facie, it might be tempting to only consider $\quantcomp$ when determining conservativity. However, the bare meanings of quantifiers can contain a dependence on $\mathbb{U}^c$, which can make the effective meaning non-conservative. Instead, we check for each quantifier in the language which set its effective meaning depends on. We consider a \textit{system} conservative if it satisfied Universal~\ref{claim:globalcons}. 

We compute which set an effective meaning depends on using the Z3 SMT solver. We first convert each effective system expressed in the operators of the LoT of Appendix~\ref{appendix:conservativity} into expressions in the Z3 SMT solver. Because every term in the LoT is a Boolean combination of $L$, $R$, and $U$, every meaning is invariant under isomorphism. Therefore, the truth value of a quantifier is a function of the cardinality of the sets $U\setminus(L\cup R)$, $L\setminus R$, $R\setminus L$, and $L\cap R$. A quantifier $Q$ is \textit{not} cons$_2$ if there is a universe $u$ and two sets $L,R \subseteq u$ such that:
\begin{equation}\label{eq:consconstraint}
    Q(L, R) \nLeftrightarrow Q(L, L\cap R)
\end{equation}
and similarly for non-cons$_2$. For each quantifier, we use the SMT solver Z3 to search for two cell-count vectors $\left(\begin{smallmatrix} |U\setminus(L\cup R)| & |L\setminus R| & |R\setminus L| & |L\cap R| \end{smallmatrix}\right)$ satisfying the constraint in Equation~\ref{eq:consconstraint}. Unsatisfiability corresponds to a conservative system, satisfiability to non-conservative. Finally, we check for each system whether all its quantifiers are uniformly cons$_1$ or cons$_2$, as stated in Universal \ref{claim:globalcons}. This procedure ensures that the systems satisfy the conservativity condition for domains of any size, and it avoids mis-classifying systems as non-conservative because of spurious appearances of, e.g., $\mathbb{U}$, such as $|\mathbb{U}| - |\mathbb{U}|$.

\end{document}